\documentclass[10pt,twocolumn,letterpaper]{article}

\usepackage[pagenumbers]{cvpr} 

\usepackage[dvipsnames]{xcolor}

\usepackage{framed}
\usepackage{algorithm}
\usepackage{algpseudocode}
\usepackage{bm}
\usepackage{setspace}
\usepackage{multirow}
\usepackage{arydshln}
\usepackage{xspace}
\usepackage{color}

\newcommand{\equalcontribution}{\textsuperscript{*}}

\title{\algoName: Towards Hetero-Client Federated Multi-Task Learning}

\author{Yuxiang Lu\equalcontribution, Suizhi Huang\equalcontribution, Yuwen Yang, Shalayiding Sirejiding, Yue Ding, Hongtao Lu\\
Department of Computer Science and Engineering, Shanghai Jiao Tong University\\
}

\newcommand{\algoName}{\textsc{FedHCA$^2$}\xspace}
\newtheorem{theorem}{Theorem}

\definecolor{cvprblue}{rgb}{0.21,0.49,0.74}
\usepackage[pagebackref,breaklinks,colorlinks,citecolor=cvprblue]{hyperref}

\begin{document}

\setlength{\abovedisplayskip}{6pt}
\setlength{\belowdisplayskip}{6pt}

\maketitle

\begin{abstract}
Federated Learning (FL) enables joint training across distributed clients using their local data privately.
Federated Multi-Task Learning (FMTL) builds on FL to handle multiple tasks, assuming model congruity that identical model architecture is deployed in each client. To relax this assumption and thus extend real-world applicability, we introduce a novel problem setting, Hetero-Client Federated Multi-Task Learning (HC-FMTL), to accommodate diverse task setups. The main challenge of HC-FMTL is the model incongruity issue that invalidates conventional aggregation methods. It also escalates the difficulties in accurate model aggregation to deal with data and task heterogeneity inherent in FMTL. To address these challenges, we propose the \algoName framework, which allows for federated training of personalized models by modeling relationships among heterogeneous clients. Drawing on our theoretical insights into the difference between multi-task and federated optimization, we propose the Hyper Conflict-Averse Aggregation scheme to mitigate conflicts during encoder updates. Additionally, inspired by task interaction in MTL, the Hyper Cross Attention Aggregation scheme uses layer-wise cross attention to enhance decoder interactions while alleviating model incongruity. Moreover, we employ learnable Hyper Aggregation Weights for each client to customize personalized parameter updates. 
Extensive experiments demonstrate the superior performance of \algoName in various HC-FMTL scenarios compared to representative methods. Our code will be made publicly available. 
\end{abstract}

\renewcommand{\thefootnote}{\fnsymbol{footnote}}
\footnotetext[1]{Equal contribution.}

\section{Introduction}
Federated Learning (FL) \cite{1stfed} has emerged as a prominent paradigm in distributed training, gaining attention in both academic and industrial fields~\cite{fedsurvey,fedsurvey2,covid,covid2,flbrain}. The FL framework empowers to collaboratively train models across multiple clients, like mobile devices or distributed data centers, while preserving data privacy and reducing communication costs. The impetus behind FL lies in the recognition that harnessing a broader dataset can improve model performance, but it also introduces the \textit{data heterogeneity} issue, as clients often collect samples from non-i.i.d. data distributions. Nevertheless, most FL research is centered on single-task scenarios, overlooking applications that demand simultaneous multi-task processing, \eg, autonomous driving~\cite{auto}. This gap has led to the integration of Multi-Task Learning (MTL) with FL, giving rise to Federated Multi-Task Learning (FMTL)~\cite{mtfl1,mtfl3}. While existing FMTL approaches primarily address statistical challenges \cite{fedmtl,fedem}, recent studies \cite{FedBone, MAS, MaT-FL} have highlighted the importance of \textit{task heterogeneity}, particularly for dense predictions such as semantic segmentation and depth estimation \cite{deeplab, dpt, pvt}.

\begin{figure}[t]
    \centering
    \includegraphics[width=0.9\linewidth]{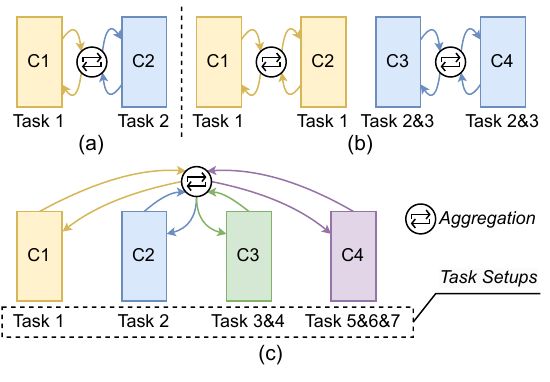}
    \vspace{-4mm}
    \caption{Comparison of different settings in FMTL. (a) Each client is dedicated to a single task. (b) Clients are grouped with peers, and peers in the same group share identical task setting. (c) Our proposed HC-FMTL setting that enables flexible collaboration among clients with different task setups.}
    \vspace{-4mm}
    \label{fig:intro}
\end{figure}

However, these FMTL methods often assume model congruity among clients, \ie, all participants either engage in a single task or aggregate with peers handling identical task sets, as shown in \cref{fig:intro}a, \ref{fig:intro}b.
Considering the discrepancy of heterogeneous tasks in practical applications as well as the expensive labor of annotating task-specific labels, clients often have different task setups in different environments.
Here \textit{task setup} describes a set of tasks that can vary in both number and type.
We define this as a new problem setting: \textbf{\textit{Hetero-Client Federated Multi-Task Learning}} (HC-FMTL), as depicted in \cref{fig:intro}c. 
HC-FMTL relaxes the constraints on model congruity, facilitating more flexible collaborative learning of various tasks across diverse private data domains and making FMTL scenarios more universally applicable.

As a more pervasive setting relaxed from FMTL, HC-FMTL introduces an additional challenge of \textbf{\textit{model incongruity}}, which exacerbates client heterogeneity. This issue arises from clients having different task setups, coupled with the prevalent use of encoder-decoder architectures in vision tasks, leading to a disparity in multi-task model structures. Model incongruity not only increases the complexity of model aggregation but also coexists with the data and task heterogeneity inherent in FMTL. Data heterogeneity is a consequence of clients encountering distinct data domains, as clients tend to use data from different domains to handle different target tasks without any overlap, which can result in performance degradation of collective learning. Meanwhile, task heterogeneity, which assigns different objectives for each task, could impede joint optimization and magnify the influence of data heterogeneity.

In this paper, we propose a novel framework named \textbf{\algoName}, designed for HC-FMTL. Our goal is to adaptively discern the relationships among heterogeneous clients and learn personalized yet globally collaborative models that benefit from both synergies and distinctions among clients and tasks. 
Since model incongruity precludes the straightforward application of conventional aggregation methods in FL, 
our approach involves the server disassembling client models into encoders and decoders for independent aggregation. For the encoders, we design the Hyper Conflict-Averse Aggregation scheme to alleviate update conflicts among clients. The motivation behind this is grounded in our theoretical analysis (see \cref{th1}) that the optimization processes of MTL and FL are closely connected and share similarities. By incorporating an approximated gradient smoothing technique, we can find an appropriate update direction for all clients that mitigates the negative effects of conflicting parameter updates caused by data and task heterogeneity. When aggregating the decoders, we devise the Hyper Cross Attention Aggregation scheme to accommodate client heterogeneity. We draw inspiration from the modeling of task interaction in MTL~\cite{cross-stitch, pad-net} and apply it to FL. Specifically, we implement a layer-wise cross attention mechanism to model the interplay between client decoders, enabling the capture of both the commonalities and discrepancies among different tasks in a fine-grained manner and thereby alleviating the incongruity at the model level.
In addition, the personalized parameter updates for each client are tailored by learnable Hyper Aggregation Weights, which encourage encoders and decoders to adaptively assimilate knowledge from peers that offer helpful complementary information.

Our contributions are summarized as follows:
\begin{itemize}[leftmargin=*,noitemsep,topsep=0pt]
    \item We introduce a novel setting of Hetero-Client Federated Multi-Task Learning (HC-FMTL) alongside the \algoName framework. It supports collaborative training across clients, each with its unique task setups, addressing the complexities of data and task heterogeneity, and the newly identified challenge of model incongruity. The relaxed setting broadens the FMTL's applicability to include a wider variety of clients, tasks, and data situations.
    \item We reveal the connection between the optimization of MTL and FL in \cref{th1} and underscore the importance of circumventing update conflicts among clients, which are exacerbated by data and task heterogeneity in HC-FMTL. We propose a Hyper Conflict-Averse Aggregation scheme, designed to alleviate the adverse effects on encoders when absorbing shared knowledge.
    \item We develop a Hyper Cross Attention Aggregation scheme to facilitate task interaction in decoders by modeling the fine-grained cross-task relationships among each decoder layer, tackling both intra- and inter-client heterogeneity. 
    \item We evaluate \algoName using a composite of two benchmark datasets, PASCAL-Context and NYUD-v2, for various HC-FMTL scenarios. Extensive experiments demonstrate that our approach outperforms existing methods.
\end{itemize}

\section{Related Work}

\subsection{Personalized Federated Learning} 
Federated Learning (FL) can be broadly classified into traditional and personalized types, depending on the characteristics of data distribution~\cite{pfedsurvey,VFL}. Traditional Federated Learning, exemplified by the widely used FedAvg~\cite{fedavg}, has undergone refinements to tackle challenges such as data heterogeneity~\cite{fedprox,fedma,fl1,fed2,pfnm,feddyn,fedgen,moon}, communication efficiency~\cite{commun1,scaffold,commun2,commun3,heterofl}, and privacy concerns~\cite{backdoor,priv1}. In contrast, personalized Federated Learning (pFL) emerges as a specialized variant designed to cater to individual client needs and address data heterogeneity more effectively~\cite{pfedsurvey,fedmtl}. Techniques like meta-learning~\cite{perfedavg}, regularization~\cite{pfedme,fedamp,ditto}, personalized-head methods~\cite{fedper,fedrep,fedrod,metasoftmax}, and other innovative approaches~\cite{fedphp,fedbn,perada} are widely employed in pFL. In essence, both traditional FL and pFL aim to grapple with the inherent challenge of data heterogeneity.

\begin{figure*}[t]
    \centering
    \includegraphics[width=0.9\linewidth]{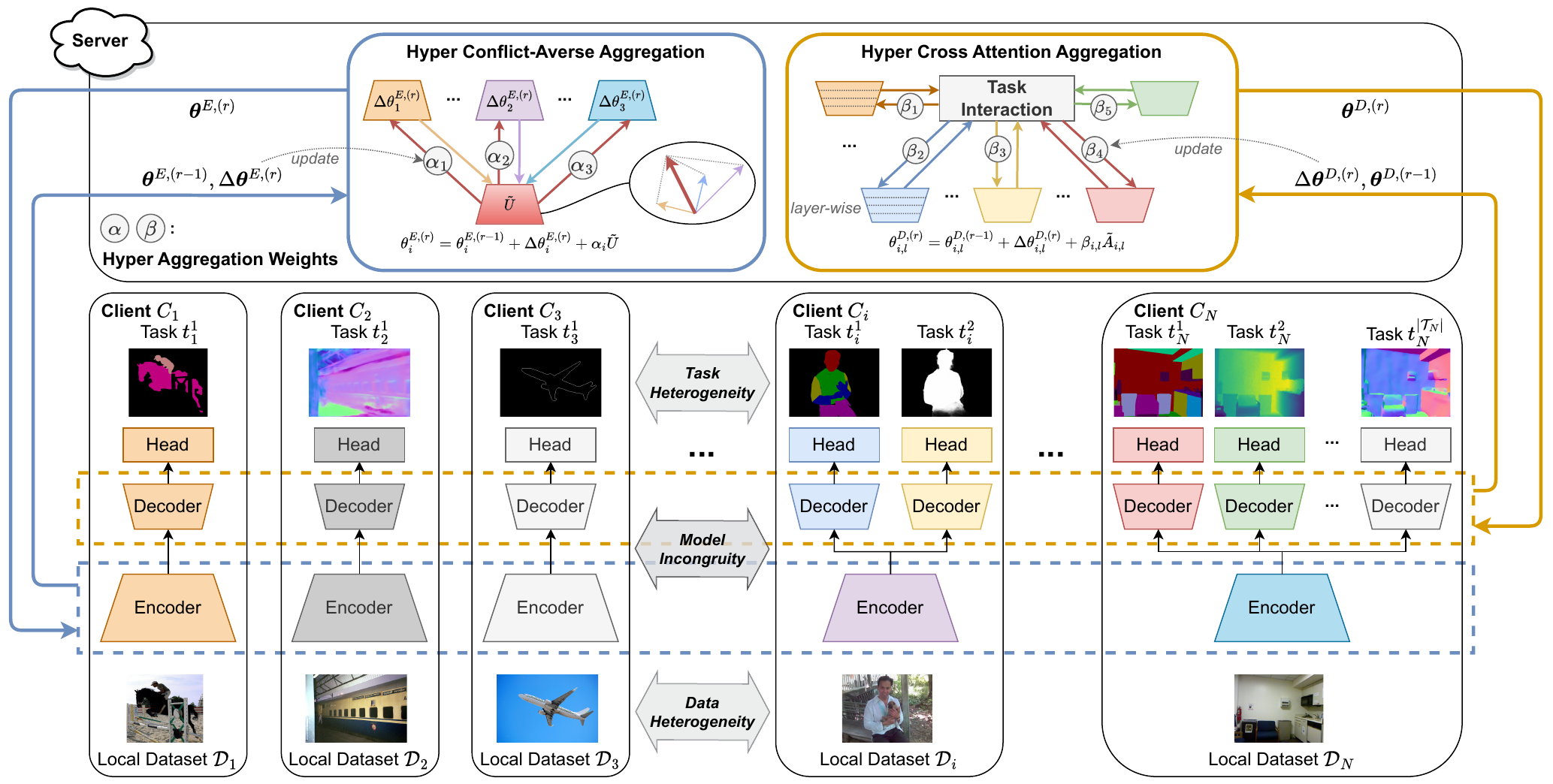}
    \vspace{-2mm}
    \caption{Illustration of the HC-FMTL setting and our proposed \algoName framework. HC-FMTL enables clients to have different task setups, from single-task (\eg client $C_1, C_2, C_3$) to multi-task (\eg client $C_i, C_N$). HC-FMTL faces three main challenges: model incongruity due to different client model structures, data heterogeneity from different local data domains, and task heterogeneity from varied target tasks. The FL system includes a server and several clients. Our framework decomposes model aggregation into two parts: Hyper Conflict-Averse Aggregation for encoders and Hyper Cross Attention Aggregation for decoders. Learnable Hyper Aggregation Weights are employed to customize personalized parameter updates and are iteratively updated by local model updates from clients.}
    \vspace{-4mm}
    \label{fig:arch}
\end{figure*}

\subsection{Multi-Task Learning}
Multi-Task Learning (MTL) aims to improve overall performance while reducing parameters and speeding up training or inference compared to training individual models for each task in isolation~\cite{mtl1, mtl2, mtl3, survey}. The main directions of MTL research can be roughly categorized into network architecture design and multi-task optimization strategy~\cite{mtl3}.
Network structure design employs methods such as parameter sharing~\cite{sluice, astmt, rcm, tsn, pgt}, task interaction~\cite{mrn, jtrl, psd, vpt1, icme, mqt}, and prediction distillation~\cite{pap-net, mti-net, invpt, invpt++}. Regarding multi-task optimization, strategies are differentiated into loss balancing and gradient balancing. Loss balancing techniques are designed to produce suitable loss weights to reduce conflicts among multiple tasks~\cite{uw,dwa,imtl,moml}. Gradient balancing, on the other hand, addresses task interference by directly adjusting gradients, with recent methods concentrating on the formulation of a unified gradient vector subject to diverse constraints~\cite{gradnorm, pcgrad, gradvac, cagrad, rotograd, graddrop, mgda}. In essence, MTL is dedicated to addressing the intrinsic challenges associated with the heterogeneity of tasks.

\subsection{Federated Multi-Task Learning}
It is essential to note that conventional Federated Multi-Task Learning (FMTL) is a branch of personalized Federated Learning that primarily deals with data heterogeneity across clients~\cite{mtfl1,mtfl2,mtfl3}. Representative works like MOCHA~\cite{fedmtl} and FedEM~\cite{fedem} attempt to train models across clients with diverse data distributions within an MTL setting.
Recent advancements, including FedBone~\cite{FedBone}, MAS~\cite{MAS}, and MaT-FL~\cite{MaT-FL}, have aimed to address both task and data heterogeneity in FMTL. FedBone aggregates the encoders by gradients uploaded from each client, enhancing feature extraction capability. MaT-FL uses dynamic grouping to combine different client models. MAS distributes varied multi-task models to clients and aggregates models among those with the same task sets.
Nevertheless, FedBone and MaT-FL are limited to each client managing a single task (\cref{fig:intro}a). MAS supports multi-task clients but is still limited to identical task sets for aggregation (\cref{fig:intro}b). In contrast, our proposed framework enables aggregation across clients with varying numbers and types of tasks, offering a more flexible collaboration.

\section{Methodology}
\subsection{Preliminary}

Within Hetero-Client Federated Multi-Task Learning (HC-FMTL), clients are assigned flexible task setups, spanning from single-task to multi-task configurations, with an arbitrary number of tasks per client. 
Formally, given a pool of $N$ clients, with client $C_i$ addressing task sets $\mathcal{T}_i$ on a corresponding local dataset $\mathcal{D}_i=\{(\mathbf{x}_n, \mathbf{y}_n)\}_{n=1}^{|\mathcal{D}_i|}$, where $\mathbf{x}_n$ is the input sample and $\mathbf{y}_n=\bigcup_{t\in \mathcal{T}_i} \mathbf{y}_{n,t}$ contains the ground-truth labels for all tasks in $\mathcal{T}_i$.

In line with FMTL, the objective of HC-FMTL is to train client-specific models ${\bm\theta}=\{\theta_1, \dots, \theta_N\}$ that benefit from collaborative optimization with other clients, thus improving performance on their local tasks.
The learning objective is to optimize personalized client models with Multi-Task Learning, formulated as follows:
\begin{equation}
    \min_{\theta_i} \sum_{t\in \mathcal{T}_i}\mathcal{L}_{i,t}(\theta_i), \quad \forall i\in \{1, \ldots, N\},
\end{equation}
where $\mathcal{L}_{i,t}$ is the loss function computed over client $C_i$'s local dataset $\mathcal{D}_i$ for task $t$.

\subsection{Architecture Overview}

The overall architecture of our proposed \algoName is depicted in \cref{fig:arch}. It contains a pool of clients that perform local training on their private datasets and a server that coordinates the aggregation of models from these clients. Concerning dense prediction tasks, each client $C_i$ utilizes an encoder-decoder structure consisting of  a shared encoder $\theta_i^E$, task-specific decoders $\{\theta_i^{D,1}, \dots, \theta_i^{D,|\mathcal{T}_i|}\}$ and prediction heads for each task type they handle. In each communication round $r$, after all clients finish their local training, they send the model parameters of previous round $\bm\theta^{(r-1)}$ and the updates in current round $\Delta\bm\theta^{(r)}$ to the server. The server first disassembles these models into encoders and decoders and then performs independent aggregation processes. The prediction heads, due to their varying parameter dimensions tailored to specific task outputs, are excluded from the aggregation process and remain localized to individual clients. The encoder parameters from all $N$ clients undergo Hyper Conflict-Averse Aggregation. Meanwhile, the server aggregates the parameters of all $K=\sum_{i=1}^{N}|\mathcal{T}_i|$ decoders through Hyper Cross Attention Aggregation.
The entire pipeline of our framework is outlined in \cref{alg1}.

\begin{algorithm}
    \small
    \setstretch{1.07}
    \renewcommand{\algorithmicrequire}{\textbf{Input:}}
	\renewcommand{\algorithmicensure}{\textbf{Output:}}
    \caption{Pseudo-codes for \algoName}
    \label{alg1}
    \begin{algorithmic}[1]
    \Require{$N$ clients $\{C_1, \dots, C_N\}$ with private local datasets $\{\mathcal{D}_1, \dots, \mathcal{D}_N\}$, client $C_i$ addresses tasks $\mathcal{T}_i$, total communication rounds $R$, local epoch $E$, learning rate $\eta$}
    \Ensure{Trained models ${\bm\theta^{(R)}}=\{\theta_1^{(R)}, \dots, \theta_N^{(R)}\}$}
    \State Clients initialize models $\bm{\theta}^{(0)}=\{\theta_1^{(0)},\dots, \theta_N^{(0)}\}$, each model $\theta_i$ consists of a shared encoder $\theta_i^E$ and $|\mathcal{T}_i|$ task-specific decoders $\bigcup_{j=1}^{|\mathcal{T}_i|}\theta_i^{D, j}$ and heads
    \State Server initializes Hyper Aggregation Weights $\bm{\alpha}$ and $\bm{\beta}$
    \Procedure{Server Update}{}
    \For{each communication round $r\in\{1, \dots, R\}$}
        \For{each client $C_i$ in parallel}
            \State $\Delta\theta_i^{(r)} \gets \Call{Client Update}{{\theta}_i^{(r-1)}}$
        \EndFor
        \State Server gathers updates of client models $\Delta\bm\theta^{(r)}$
        \State Update $\bm{\alpha}, \bm{\beta}$ using $\Delta\bm\theta^{(r)}$ with \cref{eq:hwu}
        \State ${\bm\theta}^{(r)} \gets $ \Call{Aggregation}{$\bm\theta^{(r-1)}, \Delta\bm\theta^{(r)}$}
    \EndFor
    \EndProcedure
    \Procedure{Client Update}{$\theta_i^{(r-1)}$}
    \State $\theta_i\gets \theta_i^{(r-1)}$
    \For{each local epoch $e\in\{1,\dots,E\}$}
        \For{mini-batch $\mathcal{B}_i\subset \mathcal{D}_i$}
            \State Compute losses $\mathcal{L}_i=\sum_{j=1}^{|\mathcal{T}_i|}\mathcal{L}_i^j(\theta_i;\mathcal{B}_i)$
            \State Update model $\theta_i \gets \theta_i-\eta\nabla_{\theta_i}\mathcal{L}_i$
        \EndFor
    \EndFor
    \State \textbf{return} $\Delta\theta_i^{(r)} = \theta_i - \theta_i^{(r-1)}$
    \EndProcedure
    \end{algorithmic}
\end{algorithm}

\subsection{Hyper Conflict-Averse Aggregation}
In MTL, the encoder typically employs a parameter-sharing mechanism to capture common task-agnostic information, thereby serving as a general feature extractor for all tasks and enhancing their generalization capabilities. Within our encoder aggregation process, we anticipate that encoders from various clients, each addressing distinct tasks on different data domains, are able to acquire general knowledge from other client encoders akin to MTL. To elucidate this, we begin with a theoretical analysis of the correlation between the optimization process of MTL and FL. 

\begin{figure}
    \centering
    \includegraphics[width=1.05\linewidth]{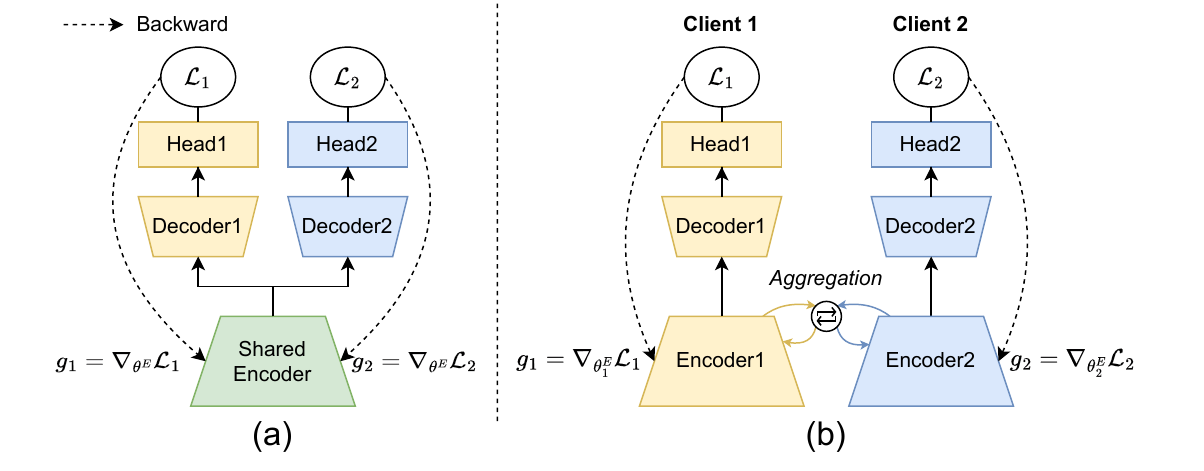}
    \vspace{-8mm}
    \caption{Comparison of optimization in MTL and FL.
    (a) The shared encoder in MTL is updated by gradient accumulation from all tasks. (b) The clients' encoders are updated independently and then aggregated in FL.}
    \vspace{-4mm}
    \label{fig:enc}
\end{figure}

As depicted in \cref{fig:enc}a, consider an MTL scenario where $N$ tasks are learned simultaneously using a standard multi-decoder architecture. In each mini-batch, the network backpropagates the loss functions $\mathcal{L}_1, \ldots, \mathcal{L}_N$ onto the shared encoder $\theta^E$ to calculate its gradient and update:
\begin{align}
    &g=\sum_{i=1}^N g_i=\sum_{i=1}^N\nabla_{\theta^E}\mathcal{L}_i,  \\
    &\Delta \theta^E=-\eta g=-\eta\sum_{i=1}^N g_i, \label{eq:mtl}
\end{align}
where $g$ represents the cumulative gradient on the encoder and $\eta$ signifies the learning rate. By updating through the summation of gradients from diverse tasks, the encoder assimilates knowledge from various task domains, aligning with the objective of MTL. Meanwhile, in an FL setting shown in \cref{fig:enc}b, suppose there are $N$ clients, each using separate networks with the same architectures as the multi-task model but with independent encoders $\theta^E_1, \dots, \theta^E_N$ to learn the same $N$ tasks. Assuming identical initial weights $\theta^E$ with MTL, and they are trained for only one mini-batch to obtain gradients $g_i=\nabla_{\theta^E}\mathcal{L}_i$. FL typically aggregates these encoders by averaging the parameters of all clients:
\begin{align}
    \tilde\theta^E_i = \frac{1}{N}\sum_{i=1}^{N}\theta^E_i,
\end{align}
where $\tilde\theta^E_i$ is the aggregated encoder parameters.
Considering its change from the initial weight:
\begin{align}
    \Delta\tilde\theta^E_i = \frac{1}{N}\sum_{i=1}^{N}\Delta\theta^E_i = \frac{1}{N}(-\eta)\sum_{i=1}^{N}g_i,
\end{align}
it means the update for the aggregated encoder mirrors the update of the shared encoder in MTL, if we regard the optimizer as capable of automatically scaling the learning rate $\eta$ in \cref{eq:mtl}.
While FL typically aggregates client models after several local training epochs in a communication round, this implies that there are differences between the learning processes of MTL and FL:
\begin{theorem}[Difference in optimizing MTL and FL]
\quad
Given clients with a shared encoder and task-specific decoder structure, the gradient descent in the shared encoder of MTL is equivalent to averaging parameter aggregation in FL, adding an extra term that maximizes the inner product of gradients between all pairs of tasks in each iteration.
\label{th1}
\end{theorem}
We provide proofs and in-depth analysis in Appendix A.
As the inner product of gradients is a measurement of accordance, maximizing the inner product is equal to reducing the conflict of gradients~\cite{mamdr}.
Hence, \cref{th1} states the necessity of integrating optimization techniques to mitigate gradient conflicts during encoder aggregation in HC-FMTL. Inspired by CAGrad~\cite{cagrad}, for each communication round, we aim to find an optimal aggregated update $\tilde U$ for the encoder that minimizes conflicts while optimizing the main objective with optimization problem:
\begin{equation}
    \max_{\tilde U}\min_i \langle \Delta \theta^E_{i},\tilde U\rangle \quad \mathrm{s.t.} \|\tilde U-\Delta\bar{\theta}^E\|\leq c \|\Delta\bar{\theta}^E\|, 
\end{equation}
where $\Delta\bar{\theta}^E=\frac{1}{N}\sum_{i=1}^{N}\Delta\theta_i^E$ is the average parameter update and $c\in[0,1)$ is a hyper-parameter controlling the convergence rate. Here $\min_i \langle \Delta \theta^E_{i},\tilde U\rangle$ measures the maximum conflict between client updates and the target update, which is an approximation to the conflict between gradients, as the server only receives parameter updates after several local training epochs rather than the gradients in each iteration. Therefore, maximizing this term can minimize the conflict in parameter optimization, which is consistent with our findings in \cref{th1}.
With constraint $\sum_{i=1}^N w_i=1, w_i\ge 0$, solving this problem using Lagrangian simplifies to:
\begin{align}
\label{cagrad}
    &\min_w F(w)=U_w^\top \Delta\bar{\theta}^E+\sqrt{\phi}\|U_w\|, \\
    &\text{where} \ U_w=\frac{1}{N}\sum_{i=1}^N w_i \Delta\theta_i^E, \phi=c^2\|\Delta\bar{\theta}^E\|^2.
\end{align}
Upon finding the optimum $w^*$ and the optimal $\lambda^{*}\,=\,\left|\left|U_{w^{*}}\right|\right|/\phi^{1/2} $, we have the unified aggregated update:
\begin{equation}
    \tilde U=\Delta\bar{\theta}^E+U_{w^{*}}/\lambda^{*}=\Delta\bar{\theta}^E+\frac{\sqrt{\phi}}{\|U_w\|}U_w. \label{eq:enc_final}
\end{equation}

\subsection{Hyper Cross Attention Aggregation}
The significance of task interaction in MTL is well-established~\cite{cross-stitch, nddr-cnn, pad-net, mti-net, atrc}, as it allows for exchanging knowledge among tasks and benefiting from complementary information. In representative methods~\cite{cross-stitch, pad-net}, task interaction is facilitated by adding the target task's feature with those from source tasks in decoders, formulated as:
\begin{equation}
    {\mathbf{z}}^D_i=\sum_{j=1}^N \gamma_{i, j} (\theta^D_j)^\top \mathbf{z}^E,\quad \forall i \in \{1, \dots, N\},
\end{equation}
where $\mathbf{z}^E$ denotes the output feature of the shared encoder, $\theta^D_j$ represents the decoder of task $j$, and $(\theta^D_j)^\top \mathbf{z}^E$ yields task-specific feature from the decoder. The coefficient $\gamma_{i, j}$ manages the flow of features from source to target tasks within the interaction and is usually a learnable parameter. To emulate this task interaction within the FL context, we intuitively aggregate the decoder parameters as follows:
\begin{align}
    \tilde{\theta}^D_i = \sum_{j=1}^N \gamma_{i, j} \ \theta^D_j.
\end{align}
Due to model incongruity in the HC-FMTL environment, the decoder parameters sent to the server originate from diverse clients with heterogeneous tasks. This intricacy leads to a complex landscape where decoders may align or diverge in both data domain and task type, requiring the aggregation process to discern the nuanced relationships among them.
Our approach improves the decoder aggregation by adopting a cross attention mechanism to further promote the exchange of inter-task knowledge among clients with model incongruity. It calculates dependencies among the local updates of $K$ decoders, thereby modeling the interplay among tasks. Recognizing that decoders often exhibit varied utilities across different network layers~\cite{bmtas, ltb, fafs, taps}, we apply a layer-wise strategy \cite{pfedla} to precisely capture the cross-task attention at each decoder layer, allows for a more fine-grained personalized aggregation that can benefit the transfer of task-specific knowledge. The computation of cross attention is defined as:
\begin{align}
    V_l &=[\Delta\theta^D_{1,l}, \dots, \Delta\theta^D_{K,l}]^\top, \\
    \tilde A_{i,l} &=\text{Softmax}({\Delta\theta^D_{i,l} \ V_l^\top}/{\sqrt{d}})V_l, \label{eq:crossatt}
\end{align}
where $[\cdot, \cdot]$ indicates concatenation, $\Delta \theta_{i,l}$ and $\tilde A_{i,l}$ are the original update and aggregated update for the $l$-th layer of the $i$-th decoder, with a dimension of $d$.

\begin{table*}[t]
    \setlength{\tabcolsep}{9pt}
    \footnotesize
    \centering
    \caption{Comparison to representative methods using PASCAL-Context for five \textbf{S}ingle-\textbf{T}ask clients and NYUD-v2 for one \textbf{M}ulti-\textbf{T}ask client. `$\uparrow$' means higher is better and `$\downarrow$' means lower is better. `$\Delta_m\%$' denotes the average performance drop w.r.t. local baseline.}
    \vspace{-3mm}
    \begin{tabular}{c|c:c:c:c:c:cccc|c}
    \hline
    \multirow{3}{*}{Method} & \multicolumn{5}{c:}{PASCAL-Context (\textbf{ST})} & \multicolumn{4}{c|}{NYUD-v2 (\textbf{MT})} & \multirow{3}{*}{${\Delta_m\%\uparrow}$} \\
    \cline{2-10}
    & SemSeg & Parts & Sal & Normals & Edge & SemSeg & Depth & Normals & Edge & \\
    & mIoU$\uparrow$ & mIoU$\uparrow$ & maxF$\uparrow$ & mErr$\downarrow$ & odsF$\uparrow$ & mIoU$\uparrow$ & RMSE$\downarrow$ & mErr$\downarrow$ & odsF$\uparrow$ & \\
    \hline
    Local & 51.69 & 49.94 & \textbf{80.91} & 15.76 & 71.95 & 41.86 & 0.6487 & 20.59 & 76.46 & 0.00 \\
    FedAvg~\cite{fedavg} & 39.98 & 37.33 & 77.56 & 18.27 & 69.17 & 38.94 & 0.7858 & 21.62 & 75.77 & -11.76 \\
    FedProx~\cite{fedprox} & 44.42 & 38.10 & 77.26 & 18.03 & 69.39 & 39.19 & 0.8068 & 21.52 & 76.03 & -10.68 \\
    FedPer~\cite{fedper} & 54.51 & 46.56 & 78.85 & 16.95 & 71.00 & \textbf{44.02} & 0.6467 & 21.19 & \textbf{76.61} & -1.11 \\
    Ditto~\cite{ditto} & 46.23 & 39.69 & 77.99 & 17.52 & 69.77 & 41.49 & 0.6508 & 20.60 & 76.45 & -5.57 \\
    FedAMP~\cite{fedamp} & 55.98 & 52.05 & 80.79 & 15.74 & 72.02 & 41.67 & 0.6428 & 20.54 & 76.40 & 1.47 \\
    MaT-FL~\cite{MaT-FL} & 57.45 & 48.63 & 79.26 & 17.26 & 71.23 & 40.99 & 0.6352 & 20.65 & 76.59 & -0.46 \\
    \algoName & \textbf{57.55} & \textbf{52.30} & 80.71 & \textbf{15.60} & \textbf{72.08} & 41.47 & \textbf{0.6281} & \textbf{20.53} & 76.50 & \textbf{2.18} \\
    \hline
    \end{tabular}
    \vspace{-3mm}
    \label{tab:res1}
\end{table*}

\subsection{Hyper Aggregation Weights}
As pointed out by pFL, a unified update for all clients is restricted in addressing client heterogeneity. 
Hence, we propose Hyper Aggregation Weights, which adaptively assess the importance of the aggregated parameters from peers and empower clients with analogous data domains and task objectives to have higher aggregation weights. This enhancement reinforces the mutual contribution from complementary information, thus serving as  high-level guidance in harmonizing the local updates with the collaborative updates. 
Specifically, the server maintains a dedicated set of weights for each client, which are applied as follows in the personalized aggregation:
\begin{equation}
    \theta_i^{(r)}=\theta_i^{(r-1)}+\Delta\theta_i^{(r)}+\psi_i \tilde\theta_i,
\end{equation}
where $\psi_i$ denotes the hyper weights for client $C_i$, \ie $\alpha_i$ for encoder or $\beta_i$ for decoder, and $\tilde\theta_i$ is the aggregated update $\tilde U$ from \cref{eq:enc_final} or $\tilde A_{i,l}$ from \cref{eq:crossatt}.
It is worth noting that we implement distinct weights for each decoder layer rather than a single weight value to be consistent with the layer-wise computation of cross attention.

\begin{table*}
    \setlength{\tabcolsep}{9pt}
    \footnotesize
    \centering
    \caption{Comparison to representative methods using NYUD-v2 for four single-task clients and PASCAL-Context for one multi-task client.}
    \vspace{-3mm}
    \begin{tabular}{c|c:c:c:c:ccccc|c}
    \hline
    \multirow{3}{*}{Method} & \multicolumn{4}{c:}{NYUD-v2 (\textbf{ST})} & \multicolumn{5}{c|}{PASCAL-Context (\textbf{MT})} & \multirow{3}{*}{${\Delta_m\%\uparrow}$} \\
    \cline{2-10}
    & SemSeg & Depth & Normals & Edge & SemSeg & Parts & Sal & Normals & Edge & \\
    & mIoU$\uparrow$ & RMSE$\downarrow$ & mErr$\downarrow$ & odsF$\uparrow$ & mIoU$\uparrow$ & mIoU$\uparrow$ & maxF$\uparrow$ & mErr$\downarrow$ & odsF$\uparrow$ & \\
    \hline
    Local & 33.59 & 0.7129 & 23.22 & 75.02 & 65.80 & 55.01 & 83.23 & 14.21 & 71.89 & 0.00 \\
    FedAvg~\cite{fedavg} & 25.80 & 0.8295 & 24.85 & 75.31 & 64.63 & 52.88 & 81.08 & 15.56 & 68.95  & -7.56 \\
    FedProx~\cite{fedprox} & 25.96 & 0.8316 & 25.20 & 75.34 & 64.97 & 50.78 & 81.29 & 15.83 &69.81 & -8.12 \\
    FedPer~\cite{fedper} & \textbf{35.93} & 0.7460 & 23.75 & \textbf{75.53} & 67.78 & 54.75 & 82.50 & 14.75 & 71.90 & -0.16  \\
    Ditto~\cite{ditto} & 28.15 & 0.7482 & 23.96 & 75.42 & 65.99 & 51.45 & 81.74 & 15.29 & 69.96 & -4.67 \\
    FedAMP~\cite{fedamp} &34.75 & 0.7103 & 23.31 &75.03 & 66.08 & 54.10 & \textbf{83.35}& 14.20 &71.88 &0.27  \\
    MaT-FL~\cite{MaT-FL} & 35.05 & 0.7504 & 23.39 & 75.33& \textbf{67.90} & 54.78 & 82.84 & 14.58 & 71.94 & -0.16 \\
    \algoName & 34.95 & \textbf{0.7018} & \textbf{23.19} & 75.03 & 65.81 & \textbf{55.01} & 83.18 & \textbf{14.08} & \textbf{71.97} & \textbf{0.75} \\
    \hline
    \end{tabular}
    \vspace{-3mm}
    \label{tab:res2}
\end{table*}

Furthermore, we design Hyper Aggregation Weights to be learnable parameters that are dynamically updated throughout the training phase. This adaptability ensures that the weights are optimized in conjunction with the system's overall objective. By employing the chain rule, we can derive the gradient of $\psi_i$ as follows:
\begin{align}
    \nabla_{\psi_i}\mathcal{L}_i = (\nabla_{\psi_i}\theta_i^{(r)})^\top\nabla_{\theta_i^{(r)}}\mathcal{L}_i = (\tilde\theta_i)^\top \nabla_{\theta_i^{(r)}}\mathcal{L}_i. \label{eq:hwg}
\end{align}
To better align this update rule with the FL paradigm, we can reformulate \cref{eq:hwg} by substituting gradients with model updates, which is the negative accumulation of gradients over batches:
\begin{align}
    \Delta \alpha_i = (\tilde U^{(r)})^\top \Delta \theta_i^{E, (r)}, \\
    \Delta \beta_{i,l} = (\tilde A^{(r)}_{i,l})^\top \Delta \theta_{i,l}^{D, (r)}.
    \label{eq:hwu}
\end{align}
It indicates that the update of Hyper Aggregation Weights can be attained by the alteration in model parameters following local training in subsequent communication rounds.

\begin{table*}[t]
    \setlength{\tabcolsep}{9pt}
    \footnotesize
    \centering
    \caption{Ablation study on our proposed aggregation schemes. `+Enc' and `+Dec' denote the integration of Hyper Conflict-Averse Aggregation for the encoders and Hyper Cross Attention Aggregation for the decoders, respectively.}
    \vspace{-3mm}
    \begin{tabular}{c|c:c:c:c:c:cccc|c}
    \hline
    \multirow{2}{*}{Method} & \multicolumn{5}{c:}{PASCAL-Context (\textbf{ST})} & \multicolumn{4}{c|}{NYUD-v2 (\textbf{MT})} & \multirow{2}{*}{${\Delta_m\%\uparrow}$} \\
    \cline{2-10}
    & SemSeg$\uparrow$ & Parts$\uparrow$ & Sal$\uparrow$ & Normals$\downarrow$ & Edge$\uparrow$ & SemSeg$\uparrow$ & Depth$\downarrow$ & Normals$\downarrow$ & Edge$\uparrow$ & \\
    \hline
    Local & 51.69 & 49.94 & \textbf{80.91} & 15.76 & 71.95 & \textbf{41.86} & 0.6487 & 20.59 & 76.46 & 0.00 \\
    +Enc & \textbf{58.38} & 51.64 & 80.44 & 15.65 & \textbf{72.09} & 41.21 & 0.6377 & 20.55 & 76.50 & 1.89 \\
    +Dec & 57.39 & 51.65 & 80.75 & 15.69 & 72.06 & 41.48 & 0.6344 & 20.56 & 76.41 & 1.80 \\
    +Enc+Dec & 57.55 & \textbf{52.30} & 80.71 & \textbf{15.60} & 72.08 & 41.47 & \textbf{0.6281} & \textbf{20.53} & \textbf{76.50} & \textbf{2.18} \\
    \hline
    \end{tabular}
    \vspace{-3mm}
    \label{tab:ab}
\end{table*}

\section{Experiments}

\subsection{Experimental Setup}
\label{setup}

\begin{figure}[t]
    \centering
    \includegraphics[width=\linewidth]{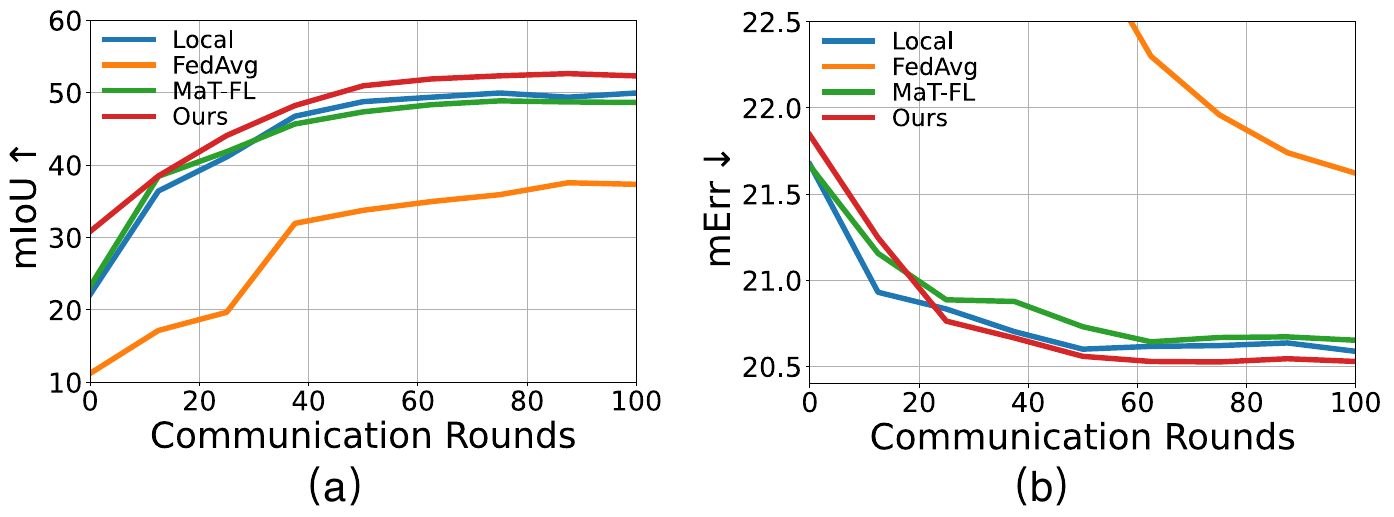}
    \vspace{-8mm}
    \caption{Evaluation results during training. (a) Parts from PASCAL-Context on single-task client. (b) Normals from NYUD-v2 on multi-task client.}
    \vspace{-4.5mm}
    \label{fig:acc}
\end{figure}

\noindent\textbf{Datasets.}
We conduct experiments with two established benchmark datasets for multi-task dense prediction: PASCAL-Context~\cite{pascal} and NYUD-v2~\cite{nyud}. The PASCAL-Context dataset contains 4,998 images for training and 5,105 for testing, annotated for five tasks: edge detection (`Edge'), semantic segmentation (`SemSeg'), human parts segmentation (`Parts'), surface normal estimation (`Normals'), and saliency detection (`Sal'). The NYUD-v2 dataset consists of 795 training images and 654 testing images, all depicting indoor scenes, and provides annotations for four tasks: edge detection, semantic segmentation, surface normal estimation, and depth estimation (`Depth').

To evaluate our algorithm, we configure two HC-FMTL benchmark scenarios: 1) Five single-task clients address five tasks in PASCAL-Context, and one multi-task client addresses four tasks in NYUD-v2; 2) Conversely, four single-task clients address four tasks in NYUD-v2, and one multi-task client addresses five tasks in PASCAL-Context. Following MaT-FL~\cite{MaT-FL}, we set an equal number of data samples among the respective clients through random partitioning.

\noindent\textbf{Implementation.} 
Our client architecture employs a pre-trained Swin-T~\cite{swin} backbone coupled with simple FCN decoders and heads. Considering the varying capacities of datasets, we use one local epoch for PASCAL-Context and four for NYUD-v2, setting the total number of communication rounds to 100 and the batch size to 8. We train all models using AdamW optimizer~\cite{adamw} with an initial learning rate and weight decay rate set at 1e-4. We implement all methods with PyTorch~\cite{pytorch} and run experiments on two NVIDIA RTX4090 GPUs. To adapt existing methods to the HC-FMTL setting, we decouple the models into encoders and decoders for separate aggregation across all methods.

\noindent\textbf{Metrics.}
We adhere to established evaluation metrics. Specifically, we measure semantic segmentation and human parts segmentation using the mean Intersection over Union (mIoU). Saliency detection is evaluated with the maximum F-measure (maxF), while surface normal estimation is assessed by the mean error (mErr). Edge detection utilizes the optimal-dataset-scale F-measure (odsF), and depth estimation uses the Root Mean Square Error (RMSE).
To provide an overall evaluation of different algorithms, we calculate the average per-task performance drop~\cite{astmt} relative to the local training baseline, which is trained without aggregation. The formula is as follows:
$ \Delta_{m}=\frac{1}{N}\sum_{i=1}^{N}(-1)^{l_{i}}\frac{M_{\texttt{Fed},i}-M_{\texttt{Local},i}}{M_{\texttt{Local},i}},$
where $N$ is the count of tasks, $M_{\texttt{Fed},i}$ and $M_{\texttt{Local},i}$ correspond to the performance of task $i$ for federated methods and the local baseline, respectively. $l_i=1$ if a lower metric value is better for task $i$, and $l_i=0$ otherwise.

\subsection{Main Results}
\label{main_res}

To evaluate the performance of our method, we compare with representative works including two traditional FL approaches FedAvg~\cite{fedavg} and FedProx~\cite{fedprox}, three pFL methods FedPer~\cite{fedper}, Ditto~\cite{ditto}, FedAMP~\cite{fedamp}, and one FMTL method MaT-FL~\cite{MaT-FL}.
The results presented in \cref{tab:res1} and \cref{tab:res2} demonstrate that \algoName consistently delivers the best performance across most metrics. More importantly, it outperforms all representative methods when considering the average per-task performance drop, which is a widely acknowledged indicator for assessing the overall performance of MTL. In addition, \cref{fig:acc} shows that \algoName converges faster to a better result on different tasks. 

\subsection{Indepth Analysis}
\noindent\textbf{Ablation Study.}
\label{ablation}
An ablation study is conducted to discern the individual contributions of each component within \algoName, as shown in~\cref{tab:ab}. The results indicate that incorporating either encoder or decoder aggregation enhances performance relative to the baseline. 
The simultaneous employment of both Hyper Conflict-Averse and Hyper Cross Attention Aggregations enables \algoName to achieve optimal performance across the evaluated configurations. This result supports the idea that using these two aggregation schemes together enhances cooperation among different clients while simultaneously reducing negative conflicts between various tasks.

\begin{table}[t]
    \setlength{\tabcolsep}{4pt}
    \footnotesize
    \centering
    \caption{Comparison between different settings. `ST+Local' and `ST+Ours' denote the setting with four single-task clients on NYUD-v2, trained with local baseline and \algoName, respectively. `ST+MT+Ours' denotes the setting in \cref{tab:res2} trained with our framework. `$\Delta_m$' is calculated w.r.t. `ST+Local' baseline.}
    \vspace{-3mm}
    \begin{tabular}{c|c:c:c:c|c}
    \hline
    Setting & SemSeg $\uparrow$ & Depth $\downarrow$ & Normals $\downarrow$ & Edge $\uparrow$ & ${\Delta_m\%\uparrow}$ \\
    \hline
    ST+Local & 33.59 & 0.7129 & 23.22 & 75.02 & 0.00 \\
    ST+Ours & 34.71 & 0.7170 & 23.25 & 74.98 & 0.64 \\
    ST+MT+Ours & \textbf{34.95} & \textbf{0.7018} & \textbf{23.19} & \textbf{75.03} & \textbf{1.44} \\
    \hline
    \end{tabular}
    \vspace{-3mm}
    \label{tab:ab2}
\end{table}

\begin{table}[t]
    \setlength{\tabcolsep}{3pt}
    \footnotesize
    \centering
    \caption{Comparison to local baseline on the setting with only multi-task clients on two datasets.}
    \vspace{-3mm}
    \begin{tabular}{c|ccc:cc|c}
    \hline
    \multirow{2}{*}{Method} & \multicolumn{3}{c:}{PASCAL-Context (\textbf{MT})} & \multicolumn{2}{c|}{NYUD-v2 (\textbf{MT})} & \multirow{2}{*}{${\Delta_m\%\uparrow}$} \\
    \cline{2-6}
    & SemSeg$\uparrow$ & Parts$\uparrow$ & Normals$\downarrow$ & SemSeg$\uparrow$& Normals$\downarrow$ & \\
    \hline
    Local & \textbf{64.87} & 53.34 & 14.07 & 39.81 & 20.65 & 0 \\
    Ours & 64.17 & \textbf{54.25} & \textbf{14.01} & \textbf{40.26} & \textbf{20.55} & \textbf{0.53} \\
    \hline
    \end{tabular}
    \vspace{-3mm}
    \label{tab:puremt}
\end{table}

\noindent\textbf{Impact of different FMTL scenarios.}
To further verify the necessity of introducing our new setting, we conduct experiments comparing two scenarios: 1) each client handles a single task, and 2) HC-FMTL encompasses both single-task and multi-task clients. As~\cref{tab:ab2} illustrates, while \algoName improves upon the local baseline in the single-task client scenario, integrating the multi-task client results in a greater enhancement. This improvement is attributed to the expanded pool of data and the knowledge jointly learned from additional tasks.
Further experiments are carried out on another scenario of HC-FMTL setting which exclusively involves multi-task clients. Specifically, we select three tasks from PASCAL-Context and two tasks from NYUD-v2 to create two multi-task client setups. The outcomes presented in \cref{tab:puremt} align with our primary findings in \cref{main_res} that nearly all metrics surpass the local baseline, further confirming the efficacy of our approach in this specialized setting.

\begin{figure}[t]
    \centering
    \includegraphics[width=0.9\linewidth]{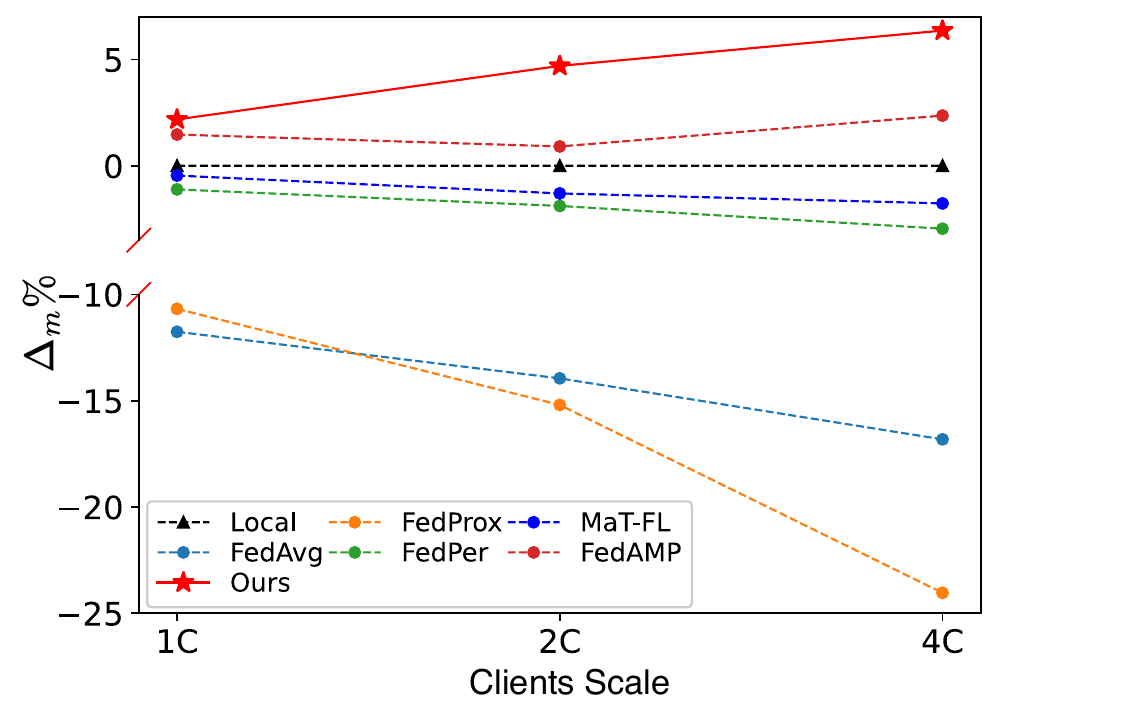}
    \vspace{-3mm}
    \caption{The performance changes of different methods with the number of clients scaling to 2 and 4 times. `$\Delta_m$' is calculated w.r.t. corresponding local baseline of 1C, 2C, or 4C. When the number of clients increases, our method can consistently provide superior performance, and an overall growth trend could be observed.}
    \vspace{-2mm}
    \label{fig:clients}
\end{figure}

\begin{figure}[t]
    \centering
    \includegraphics[width=\linewidth]{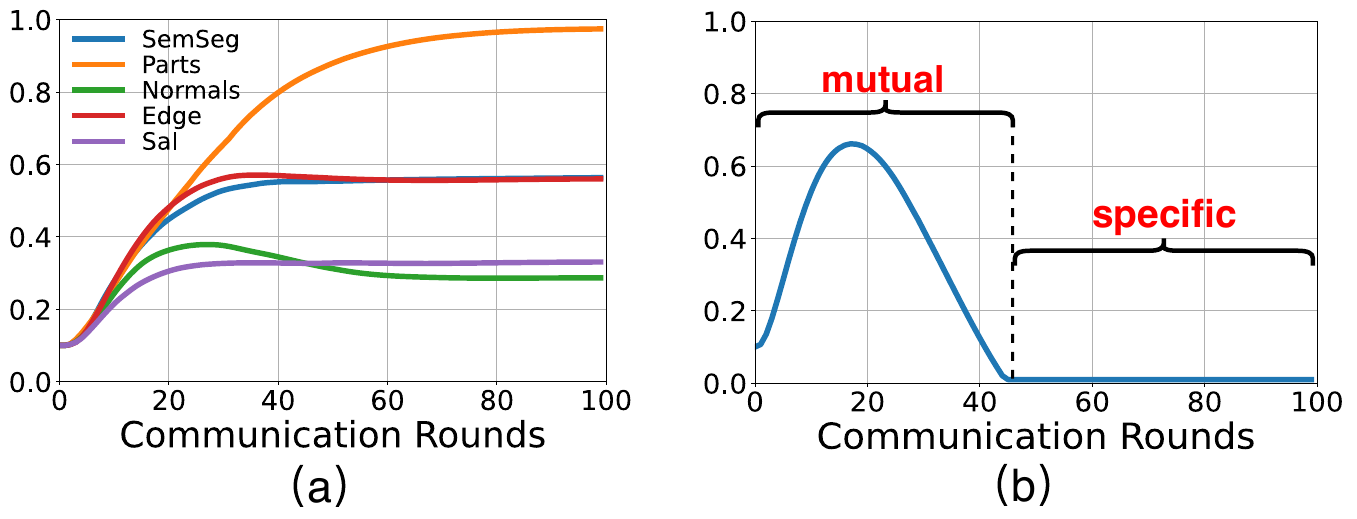}
    \vspace{-6mm}
    \caption{Hyper Aggregation Weights $\bm\alpha$ for encoders of the client models. (a) Weights of five single-task clients. (b) Weights of the multi-task client which differs in two stages.}
    \vspace{-4mm}
    \label{fig:gamma_enc}
\end{figure}

\noindent\textbf{Impact of the number of clients.}
To assess the effectiveness of \algoName across varying client counts, we conduct tests by scaling the number of clients per task by factors of 2 and 4, with the datasets evenly split.
As depicted in~\cref{fig:clients}, \algoName consistently outperforms all comparative methods, exhibiting a positive correlation between the number of clients and performance improvement. This trend contrasts with the performance decline seen with other methods as the client count increases—a result typically attributed to the diminished dataset available to each client and the increased decentralization within the federated learning system. The success of \algoName substantiates the efficacy of the Hyper Conflict-Averse Aggregation and Hyper Cross Attention Aggregation schemes, especially in scenarios characterized by pronounced data and task heterogeneity.

\begin{figure}[t]
    \centering
    \includegraphics[width=0.8\linewidth]{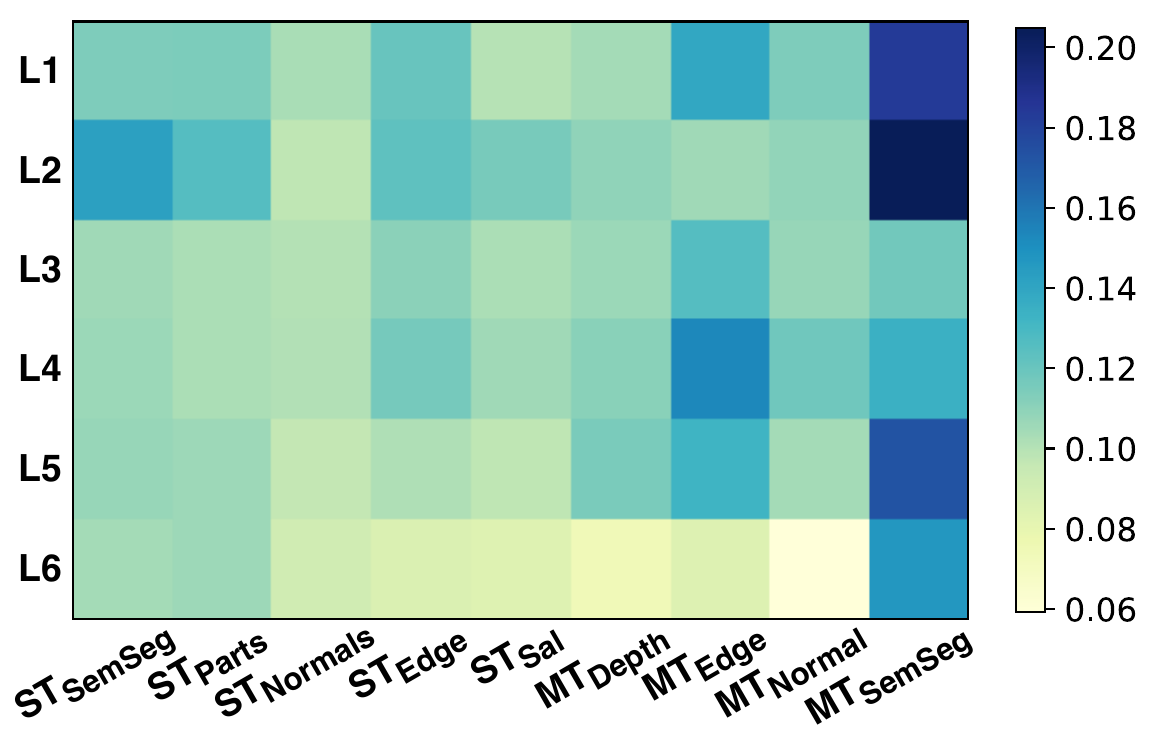}
    \vspace{-3mm}
    \caption{Learned Hyper Aggregation Weights $\bm\beta$ across decoders for different tasks, spanning layers from L1 to L6.}
    \vspace{-4mm}
    \label{fig:gamma_dec}
\end{figure}

\noindent\textbf{Interaction between tasks.}
We investigate the dynamic learning process of Hyper Aggregation Weights for both encoders and decoders, aiming to understand their role in facilitating personalized aggregation for different clients. \cref{fig:gamma_enc}a reveals that the evolution of weights for encoders in single-task clients shows a rising trend, suggesting a consistent uptake of knowledge from peers throughout the training period. In contrast, the encoder weight of the multi-task client, as depicted in \cref{fig:gamma_enc}b, exhibits two stages. Initially, the multi-task client mutually assimilates knowledge from single-task clients, a process that is crucial for rapid model convergence. The mutual learning for the multi-task client reaches its peak at about 20 rounds when the encoder weights are comparable. Subsequently, in the second phase, due to the heterogeneity in data and tasks, the multi-task client tends to enhance its feature extraction capabilities specific to its own data domain. 

Weights for decoders, as shown in~\cref{fig:gamma_dec}, vary significantly across different tasks and decoder layers. From a layer-oriented perspective, the layer closest to the output head, \ie, L6, depends least on cross-task information, which ensures that the final output is finely tuned to the specific task. In terms of task-related differences, a phenomenon markedly distinct from encoders is observed. For decoders of multi-task client, there is a persistent information integration from other tasks until the end of training. This empirical evidence substantiates the significance of employing task interaction in decoder aggregation.

\section{Conclusion}

In conclusion, this paper addresses the challenges of heterogeneity in the novel Hetero-Client Federated Multi-Task Learning (HC-FMTL) setting through the novel \algoName framework. By recognizing and tackling the issues of model incongruity, data heterogeneity, and task heterogeneity, \algoName learns personalized models with synergies of the proposed Hyper Conflict-Averse Aggregation, Hyper Cross Attention Aggregation, and Hyper Aggregation Weights. Theoretical insights and extensive experiments confirm the effectiveness of our methodology. Our work opens possibilities for more flexible FL systems in diverse and realistic settings. For future work, we aim to delve into greater model heterogeneity that accommodates varied network structures across clients, and to integrate specific modules into multi-task clients to further enhance task interaction, drawing on advancements in MTL. 

\clearpage
{
    \small
    \bibliographystyle{ieeenat_fullname}
    \bibliography{main}

\begin{thebibliography}{92}
\providecommand{\natexlab}[1]{#1}
\providecommand{\url}[1]{\texttt{#1}}
\expandafter\ifx\csname urlstyle\endcsname\relax
  \providecommand{\doi}[1]{doi: #1}\else
  \providecommand{\doi}{doi: \begingroup \urlstyle{rm}\Url}\fi

\bibitem[Acar et~al.(2021)Acar, Zhao, Matas, Mattina, Whatmough, and
  Saligrama]{feddyn}
Durmus Alp~Emre Acar, Yue Zhao, Ramon Matas, Matthew Mattina, Paul Whatmough,
  and Venkatesh Saligrama.
\newblock Federated learning based on dynamic regularization.
\newblock In \emph{ICLR}, 2021.

\bibitem[Arivazhagan et~al.(2019)Arivazhagan, Aggarwal, Singh, and
  Choudhary]{fedper}
Manoj~Ghuhan Arivazhagan, Vinay Aggarwal, Aaditya~Kumar Singh, and Sunav
  Choudhary.
\newblock Federated learning with personalization layers.
\newblock \emph{CoRR}, abs/1912.00818, 2019.

\bibitem[Bagdasaryan et~al.(2020)Bagdasaryan, Veit, Hua, Estrin, and
  Shmatikov]{backdoor}
Eugene Bagdasaryan, Andreas Veit, Yiqing Hua, Deborah Estrin, and Vitaly
  Shmatikov.
\newblock How to backdoor federated learning.
\newblock In \emph{AISTATS}, pages 2938--2948, 2020.

\bibitem[Bai et~al.(2021)Bai, Wang, Ma, Xu, et~al.]{covid}
Xiang Bai, Hanchen Wang, Liya Ma, Yongchao Xu, et~al.
\newblock Advancing {{COVID-19}} diagnosis with privacy-preserving
  collaboration in artificial intelligence.
\newblock \emph{Nature Machine Intelligence}, 3\penalty0 (12):\penalty0
  1081--1089, 2021.

\bibitem[Bercea et~al.(2022)Bercea, Wiestler, Rueckert, and
  Albarqouni]{flbrain}
Cosmin~I. Bercea, Benedikt Wiestler, Daniel Rueckert, and Shadi Albarqouni.
\newblock Federated disentangled representation learning for unsupervised brain
  anomaly detection.
\newblock \emph{Nature Machine Intelligence}, 4\penalty0 (8):\penalty0
  685--695, 2022.

\bibitem[Brüggemann et~al.(2020)Brüggemann, Kanakis, Georgoulis, and
  Van~Gool]{bmtas}
David Brüggemann, Menelaos Kanakis, Stamatios Georgoulis, and Luc Van~Gool.
\newblock Automated search for resource-efficient branched multi-task networks.
\newblock In \emph{BMVC}, page 359, 2020.

\bibitem[Brüggemann et~al.(2021)Brüggemann, Kanakis, Obukhov, Georgoulis, and
  Van~Gool]{atrc}
David Brüggemann, Menelaos Kanakis, Anton Obukhov, Stamatios Georgoulis, and
  Luc Van~Gool.
\newblock Exploring relational context for multi-task dense prediction.
\newblock In \emph{ICCV}, pages 15869--15878, 2021.

\bibitem[Cai et~al.(2023)Cai, Chen, Liu, Srinivasa, Lee, Kompella, and
  Wang]{MaT-FL}
Ruisi Cai, Xiaohan Chen, Shiwei Liu, Jayanth Srinivasa, Myungjin Lee, Ramana
  Kompella, and Zhangyang Wang.
\newblock Many-task federated learning: A new problem setting and a simple
  baseline.
\newblock In \emph{CVPR}, pages 5037--5045, 2023.

\bibitem[Caruana(1997)]{mtl1}
Rich Caruana.
\newblock Multitask learning.
\newblock \emph{Machine learning}, 28\penalty0 (1):\penalty0 41--75, 1997.

\bibitem[Chen and Chao(2022)]{fedrod}
Hong-You Chen and Wei-Lun Chao.
\newblock On bridging generic and personalized federated learning for image
  classification.
\newblock In \emph{ICLR}, 2022.

\bibitem[Chen et~al.(2018{\natexlab{a}})Chen, Zhu, Papandreou, Schroff, and
  Adam]{deeplab}
Liang-Chieh Chen, Yukun Zhu, George Papandreou, Florian Schroff, and Hartwig
  Adam.
\newblock Encoder-decoder with atrous separable convolution for semantic image
  segmentation.
\newblock In \emph{ECCV}, pages 801--818, 2018{\natexlab{a}}.

\bibitem[Chen et~al.(2023)Chen, Zhang, Jiang, Chen, Gao, and Huang]{FedBone}
Yiqiang Chen, Teng Zhang, Xinlong Jiang, Qian Chen, Chenlong Gao, and Wuliang
  Huang.
\newblock Fedbone: Towards large-scale federated multi-task learning.
\newblock \emph{arXiv preprint arXiv:2306.17465}, 2023.

\bibitem[Chen et~al.(2018{\natexlab{b}})Chen, Badrinarayanan, Lee, and
  Rabinovich]{gradnorm}
Zhao Chen, Vijay Badrinarayanan, Chen-Yu Lee, and Andrew Rabinovich.
\newblock Gradnorm: Gradient normalization for adaptive loss balancing in deep
  multitask networks.
\newblock In \emph{ICML}, pages 794--803, 2018{\natexlab{b}}.

\bibitem[Chen et~al.(2020)Chen, Ngiam, Huang, Luong, Kretzschmar, Chai, and
  Anguelov]{graddrop}
Zhao Chen, Jiquan Ngiam, Yanping Huang, Thang Luong, Henrik Kretzschmar, Yuning
  Chai, and Dragomir Anguelov.
\newblock Just pick a sign: Optimizing deep multitask models with gradient sign
  dropout.
\newblock In \emph{NeurIPS}, 2020.

\bibitem[Collins et~al.(2021)Collins, Hassani, Mokhtari, and
  Shakkottai]{fedrep}
Liam Collins, Hamed Hassani, Aryan Mokhtari, and Sanjay Shakkottai.
\newblock Exploiting shared representations for personalized federated
  learning.
\newblock In \emph{ICML}, pages 2089--2099, 2021.

\bibitem[Crawshaw(2020)]{mtl3}
Michael Crawshaw.
\newblock Multi-task learning with deep neural networks: A survey.
\newblock \emph{arXiv preprint arXiv:2009.09796}, 2020.

\bibitem[Dayan et~al.(2021)Dayan, Roth, Zhong, Harouni, Gentili,
  et~al.]{covid2}
Ittai Dayan, Holger~R. Roth, Aoxiao Zhong, Ahmed Harouni, Amilcare Gentili,
  et~al.
\newblock Federated learning for predicting clinical outcomes in patients with
  {{COVID-19}}.
\newblock \emph{Nature Medicine}, 27\penalty0 (10):\penalty0 1735--1743, 2021.

\bibitem[Diao et~al.(2021)Diao, Ding, and Tarokh]{heterofl}
Enmao Diao, Jie Ding, and Vahid Tarokh.
\newblock Heterofl: Computation and communication efficient federated learning
  for heterogeneous clients.
\newblock In \emph{ECCV}, 2021.

\bibitem[Fallah et~al.(2020)Fallah, Mokhtari, and Ozdaglar]{perfedavg}
Alireza Fallah, Aryan Mokhtari, and Asuman Ozdaglar.
\newblock Personalized federated learning with theoretical guarantees: A
  model-agnostic meta-learning approach.
\newblock \emph{NeurIPS}, 33:\penalty0 3557--3568, 2020.

\bibitem[Gao et~al.(2019)Gao, Ma, Zhao, Liu, and Yuille]{nddr-cnn}
Yuan Gao, Jiayi Ma, Mingbo Zhao, Wei Liu, and Alan~L Yuille.
\newblock Nddr-cnn: Layerwise feature fusing in multi-task cnns by neural
  discriminative dimensionality reduction.
\newblock In \emph{CVPR}, pages 3205--3214, 2019.

\bibitem[Guo et~al.(2020)Guo, Lee, and Ulbricht]{ltb}
Pengsheng Guo, Chen-Yu Lee, and Daniel Ulbricht.
\newblock Learning to branch for multi-task learning.
\newblock In \emph{ICML}, pages 3854--3863, 2020.

\bibitem[He et~al.(2022)He, Ceyani, Balasubramanian, Annavaram, and
  Avestimehr]{mtfl3}
Chaoyang He, Emir Ceyani, Keshav Balasubramanian, Murali Annavaram, and Salman
  Avestimehr.
\newblock Spreadgnn: Decentralized multi-task federated learning for graph
  neural networks on molecular data.
\newblock In \emph{AAAI}, pages 6865--6873, 2022.

\bibitem[Hu et~al.(2023)Hu, Yang, Chen, Li, Sima, Zhu, Chai, Du, Lin, Wang,
  et~al.]{auto}
Yihan Hu, Jiazhi Yang, Li Chen, Keyu Li, Chonghao Sima, Xizhou Zhu, Siqi Chai,
  Senyao Du, Tianwei Lin, Wenhai Wang, et~al.
\newblock Planning-oriented autonomous driving.
\newblock In \emph{CVPR}, pages 17853--17862, 2023.

\bibitem[Huang et~al.(2021{\natexlab{a}})Huang, Chu, Zhou, Wang, Liu, Pei, and
  Zhang]{fedamp}
Yutao Huang, Lingyang Chu, Zirui Zhou, Lanjun Wang, Jiangchuan Liu, Jian Pei,
  and Yong Zhang.
\newblock Personalized cross-silo federated learning on non-iid data.
\newblock In \emph{AAAI}, pages 7865--7873, 2021{\natexlab{a}}.

\bibitem[Huang et~al.(2021{\natexlab{b}})Huang, Gupta, Song, Li, and
  Arora]{priv1}
Yangsibo Huang, Samyak Gupta, Zhao Song, Kai Li, and Sanjeev Arora.
\newblock Evaluating gradient inversion attacks and defenses in federated
  learning.
\newblock \emph{NeurIPS}, 34:\penalty0 7232--7241, 2021{\natexlab{b}}.

\bibitem[Javaloy and Valera(2022)]{rotograd}
Adri{\'{a}}n Javaloy and Isabel Valera.
\newblock Rotograd: Gradient homogenization in multitask learning.
\newblock In \emph{ICLR}, 2022.

\bibitem[Kairouz et~al.(2021)Kairouz, McMahan, Avent, et~al.]{fedsurvey2}
Peter Kairouz, H.~Brendan McMahan, Brendan Avent, et~al.
\newblock Advances and open problems in federated learning.
\newblock \emph{Found. Trends Mach. Learn.}, 14\penalty0 (1-2):\penalty0
  1--210, 2021.

\bibitem[Kanakis et~al.(2020)Kanakis, Bruggemann, Saha, Georgoulis, Obukhov,
  and Gool]{rcm}
Menelaos Kanakis, David Bruggemann, Suman Saha, Stamatios Georgoulis, Anton
  Obukhov, and Luc~Van Gool.
\newblock Reparameterizing convolutions for incremental multi-task learning
  without task interference.
\newblock In \emph{ECCV}, pages 689--707, 2020.

\bibitem[Karimireddy et~al.(2020)Karimireddy, Kale, Mohri, Reddi, Stich, and
  Suresh]{scaffold}
Sai~Praneeth Karimireddy, Satyen Kale, Mehryar Mohri, Sashank~J. Reddi,
  Sebastian~U. Stich, and Ananda~Theertha Suresh.
\newblock {SCAFFOLD:} stochastic controlled averaging for federated learning.
\newblock In \emph{{ICML}}, pages 5132--5143, 2020.

\bibitem[Kendall et~al.(2018)Kendall, Gal, and Cipolla]{uw}
Alex Kendall, Yarin Gal, and Roberto Cipolla.
\newblock Multi-task learning using uncertainty to weigh losses for scene
  geometry and semantics.
\newblock In \emph{CVPR}, pages 7482--7491, 2018.

\bibitem[Kone{\v{c}}n{\'y} et~al.(2015)Kone{\v{c}}n{\'y}, McMahan, and
  Ramage]{1stfed}
Jakub Kone{\v{c}}n{\'y}, Brendan McMahan, and Daniel Ramage.
\newblock Federated optimization: Distributed optimization beyond the
  datacenter.
\newblock \emph{CoRR}, abs/1511.03575, 2015.

\bibitem[Kone{\v{c}}n{\'y} et~al.(2016)Kone{\v{c}}n{\'y}, McMahan, Yu,
  Richt{\'{a}}rik, Suresh, and Bacon]{commun2}
Jakub Kone{\v{c}}n{\'y}, H.~Brendan McMahan, Felix~X. Yu, Peter
  Richt{\'{a}}rik, Ananda~Theertha Suresh, and Dave Bacon.
\newblock Federated learning: Strategies for improving communication
  efficiency.
\newblock \emph{CoRR}, abs/1610.05492, 2016.

\bibitem[Li et~al.(2021{\natexlab{a}})Li, He, and Song]{moon}
Qinbin Li, Bingsheng He, and Dawn Song.
\newblock Model-contrastive federated learning.
\newblock In \emph{CVPR}, pages 10713--10722, 2021{\natexlab{a}}.

\bibitem[Li et~al.(2020)Li, Sahu, Zaheer, Sanjabi, Talwalkar, and
  Smith]{fedprox}
Tian Li, Anit~Kumar Sahu, Manzil Zaheer, Maziar Sanjabi, Ameet Talwalkar, and
  Virginia Smith.
\newblock Federated optimization in heterogeneous networks.
\newblock In \emph{MLSys}, 2020.

\bibitem[Li et~al.(2021{\natexlab{b}})Li, Hu, Beirami, and Smith]{ditto}
Tian Li, Shengyuan Hu, Ahmad Beirami, and Virginia Smith.
\newblock Ditto: Fair and robust federated learning through personalization.
\newblock In \emph{ICML}, pages 6357--6368, 2021{\natexlab{b}}.

\bibitem[Li et~al.(2021{\natexlab{c}})Li, JIANG, Zhang, Kamp, and Dou]{fedbn}
Xiaoxiao Li, Meirui JIANG, Xiaofei Zhang, Michael Kamp, and Qi Dou.
\newblock Fed{BN}: Federated learning on non-{IID} features via local batch
  normalization.
\newblock In \emph{ICLR}, 2021{\natexlab{c}}.

\bibitem[Li et~al.(2021{\natexlab{d}})Li, Zhan, Shao, Li, and Song]{fedphp}
Xin-Chun Li, De-Chuan Zhan, Yunfeng Shao, Bingshuai Li, and Shaoming Song.
\newblock Fedphp: Federated personalization with inherited private models.
\newblock In \emph{ECML PKDD}, pages 587--602, 2021{\natexlab{d}}.

\bibitem[Liu et~al.(2021{\natexlab{a}})Liu, Liu, Jin, Stone, and Liu]{cagrad}
Bo Liu, Xingchao Liu, Xiaojie Jin, Peter Stone, and Qiang Liu.
\newblock Conflict-averse gradient descent for multi-task learning.
\newblock In \emph{NeurIPS}, pages 18878--18890, 2021{\natexlab{a}}.

\bibitem[Liu et~al.(2022{\natexlab{a}})Liu, Hu, Wu, and Smith]{mtfl2}
Ken Liu, Shengyuan Hu, Steven~Z Wu, and Virginia Smith.
\newblock On privacy and personalization in cross-silo federated learning.
\newblock \emph{NeurIPS}, 35:\penalty0 5925--5940, 2022{\natexlab{a}}.

\bibitem[Liu et~al.(2021{\natexlab{b}})Liu, Li, Kuang, Xue, Chen, Yang, Liao,
  and Zhang]{imtl}
Liyang Liu, Yi Li, Zhanghui Kuang, Jing{-}Hao Xue, Yimin Chen, Wenming Yang,
  Qingmin Liao, and Wayne Zhang.
\newblock Towards impartial multi-task learning.
\newblock In \emph{ICLR}, 2021{\natexlab{b}}.

\bibitem[Liu et~al.(2019)Liu, Johns, and Davison]{dwa}
Shikun Liu, Edward Johns, and Andrew~J. Davison.
\newblock End-to-end multi-task learning with attention.
\newblock In \emph{CVPR}, pages 1871--1880, 2019.

\bibitem[Liu et~al.(2022{\natexlab{b}})Liu, Kang, Zou, Pu, He, Ye, Ouyang,
  Zhang, and Yang]{VFL}
Yang Liu, Yan Kang, Tianyuan Zou, Yanhong Pu, Yuanqin He, Xiaozhou Ye, Ye
  Ouyang, Ya{-}Qin Zhang, and Qiang Yang.
\newblock Vertical federated learning.
\newblock \emph{CoRR}, abs/2211.12814, 2022{\natexlab{b}}.

\bibitem[Liu et~al.(2021{\natexlab{c}})Liu, Lin, Cao, Hu, Wei, Zhang, Lin, and
  Guo]{swin}
Ze Liu, Yutong Lin, Yue Cao, Han Hu, Yixuan Wei, Zheng Zhang, Stephen Lin, and
  Baining Guo.
\newblock Swin transformer: Hierarchical vision transformer using shifted
  windows.
\newblock In \emph{ICCV}, pages 10012--10022, 2021{\natexlab{c}}.

\bibitem[Long et~al.(2017)Long, Cao, Wang, and Yu]{mrn}
Mingsheng Long, Zhangjie Cao, Jianmin Wang, and Philip~S Yu.
\newblock Learning multiple tasks with multilinear relationship networks.
\newblock \emph{NeurIPS}, 30, 2017.

\bibitem[Loshchilov and Hutter(2019)]{adamw}
Ilya Loshchilov and Frank Hutter.
\newblock Decoupled weight decay regularization.
\newblock In \emph{ICLR}, 2019.

\bibitem[Lu et~al.(2017)Lu, Kumar, Zhai, Cheng, Javidi, and Feris]{fafs}
Yongxi Lu, Abhishek Kumar, Shuangfei Zhai, Yu Cheng, Tara Javidi, and Rogerio
  Feris.
\newblock Fully-adaptive feature sharing in multi-task networks with
  applications in person attribute classification.
\newblock In \emph{CVPR}, pages 5334--5343, 2017.

\bibitem[Lu et~al.(2023)Lu, Sirejiding, Ding, Wang, and Lu]{pgt}
Yuxiang Lu, Shalayiding Sirejiding, Yue Ding, Chunlin Wang, and Hongtao Lu.
\newblock Prompt guided transformer for multi-task dense prediction.
\newblock \emph{arXiv preprint arXiv:2307.15362}, 2023.

\bibitem[Luo et~al.(2023)Luo, Li, Gao, Tang, Wang, Li, Zhu, Liu, Li, and
  Pan]{mamdr}
Linhao Luo, Yumeng Li, Buyu Gao, Shuai Tang, Sinan Wang, Jiancheng Li, Tanchao
  Zhu, Jiancai Liu, Zhao Li, and Shirui Pan.
\newblock {MAMDR:} {A} model agnostic learning framework for multi-domain
  recommendation.
\newblock In \emph{ICDE}, pages 3079--3092, 2023.

\bibitem[Ma et~al.(2022)Ma, Zhang, Guo, and Xu]{pfedla}
Xiaosong Ma, Jie Zhang, Song Guo, and Wenchao Xu.
\newblock Layer-wised model aggregation for personalized federated learning.
\newblock In \emph{CVPR}, pages 10092--10101, 2022.

\bibitem[Maninis et~al.(2019)Maninis, Radosavovic, and Kokkinos]{astmt}
Kevis-Kokitsi Maninis, Ilija Radosavovic, and Iasonas Kokkinos.
\newblock Attentive single-tasking of multiple tasks.
\newblock In \emph{CVPR}, pages 1851--1860, 2019.

\bibitem[Marfoq et~al.(2021)Marfoq, Neglia, Bellet, Kameni, and Vidal]{fedem}
Othmane Marfoq, Giovanni Neglia, Aur{\'{e}}lien Bellet, Laetitia Kameni, and
  Richard Vidal.
\newblock Federated multi-task learning under a mixture of distributions.
\newblock In \emph{NeurIPS}, pages 15434--15447, 2021.

\bibitem[McMahan et~al.(2017{\natexlab{a}})McMahan, Moore, Ramage, Hampson, and
  y~Arcas]{commun1}
Brendan McMahan, Eider Moore, Daniel Ramage, Seth Hampson, and Blaise~Aguera y
  Arcas.
\newblock Communication-efficient learning of deep networks from decentralized
  data.
\newblock In \emph{AISTATS}, pages 1273--1282, 2017{\natexlab{a}}.

\bibitem[McMahan et~al.(2017{\natexlab{b}})McMahan, Moore, Ramage,
  et~al.]{fedavg}
Brendan McMahan, Eider Moore, Daniel Ramage, et~al.
\newblock Communication-efficient learning of deep networks from decentralized
  data.
\newblock In \emph{AISTATS}, pages 1273--1282, 2017{\natexlab{b}}.

\bibitem[Mills et~al.(2021)Mills, Hu, and Min]{mtfl1}
Jed Mills, Jia Hu, and Geyong Min.
\newblock Multi-task federated learning for personalised deep neural networks
  in edge computing.
\newblock \emph{TPDS}, 33\penalty0 (3):\penalty0 630--641, 2021.

\bibitem[Misra et~al.(2016)Misra, Shrivastava, Gupta, and Hebert]{cross-stitch}
Ishan Misra, Abhinav Shrivastava, Abhinav Gupta, and Martial Hebert.
\newblock Cross-stitch networks for multi-task learning.
\newblock In \emph{CVPR}, pages 3994--4003, 2016.

\bibitem[Mottaghi et~al.(2014)Mottaghi, Chen, Liu, Cho, Lee, Fidler, Urtasun,
  and Yuille]{pascal}
Roozbeh Mottaghi, Xianjie Chen, Xiaobai Liu, Nam-Gyu Cho, Seong-Whan Lee, Sanja
  Fidler, Raquel Urtasun, and Alan Yuille.
\newblock The role of context for object detection and semantic segmentation in
  the wild.
\newblock In \emph{CVPR}, pages 891--898, 2014.

\bibitem[Paszke et~al.(2019)Paszke, Gross, Massa, Lerer, Bradbury, Chanan,
  Killeen, Lin, Gimelshein, and Antiga]{pytorch}
Adam Paszke, Sam Gross, Francisco Massa, Adam Lerer, James Bradbury, Gregory
  Chanan, Trevor Killeen, Zeming Lin, Natalia Gimelshein, and Luca Antiga.
\newblock Pytorch: An imperative style, high-performance deep learning library.
\newblock \emph{NeurIPS}, 32, 2019.

\bibitem[Ranftl et~al.(2021)Ranftl, Bochkovskiy, and Koltun]{dpt}
René Ranftl, Alexey Bochkovskiy, and Vladlen Koltun.
\newblock Vision transformers for dense prediction.
\newblock In \emph{ICCV}, pages 12179--12188, 2021.

\bibitem[Ren et~al.(2020)Ren, Yu, Ma, Zhao, Yi, et~al.]{metasoftmax}
Jiawei Ren, Cunjun Yu, Xiao Ma, Haiyu Zhao, Shuai Yi, et~al.
\newblock Balanced meta-softmax for long-tailed visual recognition.
\newblock \emph{NeurIPS}, 33:\penalty0 4175--4186, 2020.

\bibitem[Ruder(2017)]{mtl2}
Sebastian Ruder.
\newblock An overview of multi-task learning in deep neural networks.
\newblock \emph{arXiv preprint arXiv:1706.05098}, 2017.

\bibitem[Ruder et~al.(2019)Ruder, Bingel, Augenstein, and Søgaard]{sluice}
Sebastian Ruder, Joachim Bingel, Isabelle Augenstein, and Anders Søgaard.
\newblock Latent multi-task architecture learning.
\newblock In \emph{AAAI}, pages 4822--4829, 2019.

\bibitem[Sener and Koltun(2018)]{mgda}
Ozan Sener and Vladlen Koltun.
\newblock Multi-task learning as multi-objective optimization.
\newblock In \emph{NeurIPS}, pages 525--536, 2018.

\bibitem[Silberman et~al.(2012)Silberman, Hoiem, Kohli, and Fergus]{nyud}
Nathan Silberman, Derek Hoiem, Pushmeet Kohli, and Rob Fergus.
\newblock Indoor segmentation and support inference from rgbd images.
\newblock In \emph{ECCV}, pages 746--760, 2012.

\bibitem[Sirejiding et~al.(2023)Sirejiding, Lu, Lu, and Ding]{icme}
Shalayiding Sirejiding, Yuxiang Lu, Hongtao Lu, and Yue Ding.
\newblock Scale-aware task message transferring for multi-task learning.
\newblock In \emph{ICME}, pages 1859--1864, 2023.

\bibitem[Smith et~al.(2017)Smith, Chiang, Sanjabi, and Talwalkar]{fedmtl}
Virginia Smith, Chao{-}Kai Chiang, Maziar Sanjabi, and Ameet Talwalkar.
\newblock Federated multi-task learning.
\newblock In \emph{NeurIPS}, pages 4424--4434, 2017.

\bibitem[Sun et~al.(2021)Sun, Probst, Paudel, Popović, Kanakis, Patel, Dai,
  and Van~Gool]{tsn}
Guolei Sun, Thomas Probst, Danda~Pani Paudel, Nikola Popović, Menelaos
  Kanakis, Jagruti Patel, Dengxin Dai, and Luc Van~Gool.
\newblock Task switching network for multi-task learning.
\newblock In \emph{ICCV}, pages 8291--8300, 2021.

\bibitem[T~Dinh et~al.(2020)T~Dinh, Tran, and Nguyen]{pfedme}
Canh T~Dinh, Nguyen Tran, and Josh Nguyen.
\newblock Personalized federated learning with moreau envelopes.
\newblock \emph{NeurIPS}, 33:\penalty0 21394--21405, 2020.

\bibitem[Tan et~al.(2022)Tan, Yu, Cui, and Yang]{pfedsurvey}
Alysa~Ziying Tan, Han Yu, Lizhen Cui, and Qiang Yang.
\newblock Towards {{Personalized Federated Learning}}.
\newblock \emph{IEEE Transactions on Neural Networks and Learning Systems},
  pages 1--17, 2022.

\bibitem[Vandenhende et~al.(2020)Vandenhende, Georgoulis, and Gool]{mti-net}
Simon Vandenhende, Stamatios Georgoulis, and Luc~Van Gool.
\newblock Mti-net: Multi-scale task interaction networks for multi-task
  learning.
\newblock In \emph{ECCV}, pages 527--543, 2020.

\bibitem[Vandenhende et~al.(2021)Vandenhende, Georgoulis, Van~Gansbeke,
  Proesmans, Dai, and Van~Gool]{survey}
Simon Vandenhende, Stamatios Georgoulis, Wouter Van~Gansbeke, Marc Proesmans,
  Dengxin Dai, and Luc Van~Gool.
\newblock Multi-task learning for dense prediction tasks: A survey.
\newblock \emph{IEEE TPAMI}, 44\penalty0 (7):\penalty0 3614--3633, 2021.

\bibitem[Wallingford et~al.(2022)Wallingford, Li, Achille, Ravichandran,
  Fowlkes, Bhotika, and Soatto]{taps}
Matthew Wallingford, Hao Li, Alessandro Achille, Avinash Ravichandran, Charless
  Fowlkes, Rahul Bhotika, and Stefano Soatto.
\newblock Task adaptive parameter sharing for multi-task learning.
\newblock In \emph{CVPR}, pages 7561--7570, 2022.

\bibitem[Wang et~al.(2020{\natexlab{a}})Wang, Yurochkin, Sun, Papailiopoulos,
  and Khazaeni]{fedma}
Hongyi Wang, Mikhail Yurochkin, Yuekai Sun, Dimitris~S. Papailiopoulos, and
  Yasaman Khazaeni.
\newblock Federated learning with matched averaging.
\newblock In \emph{ECCV}, 2020{\natexlab{a}}.

\bibitem[Wang et~al.(2020{\natexlab{b}})Wang, Liu, Liang, Joshi, and Poor]{fl1}
Jianyu Wang, Qinghua Liu, Hao Liang, Gauri Joshi, and H.~Vincent Poor.
\newblock Tackling the objective inconsistency problem in heterogeneous
  federated optimization.
\newblock In \emph{NeurIPS}, 2020{\natexlab{b}}.

\bibitem[Wang et~al.(2021{\natexlab{a}})Wang, Xie, Li, Fan, Song, Liang, Lu,
  Luo, and Shao]{pvt}
Wenhai Wang, Enze Xie, Xiang Li, Deng-Ping Fan, Kaitao Song, Ding Liang, Tong
  Lu, Ping Luo, and Ling Shao.
\newblock Pyramid vision transformer: A versatile backbone for dense prediction
  without convolutions.
\newblock In \emph{ICCV}, pages 568--578, 2021{\natexlab{a}}.

\bibitem[Wang et~al.(2021{\natexlab{b}})Wang, Tsvetkov, Firat, and
  Cao]{gradvac}
Zirui Wang, Yulia Tsvetkov, Orhan Firat, and Yuan Cao.
\newblock Gradient vaccine: Investigating and improving multi-task optimization
  in massively multilingual models.
\newblock In \emph{ICLR}, 2021{\natexlab{b}}.

\bibitem[Xie et~al.(2023)Xie, Huang, Chu, Xu, Xiao, Li, and Anandkumar]{perada}
Chulin Xie, De{-}An Huang, Wenda Chu, Daguang Xu, Chaowei Xiao, Bo Li, and
  Anima Anandkumar.
\newblock Perada: Parameter-efficient and generalizable federated learning
  personalization with guarantees.
\newblock \emph{CoRR}, abs/2302.06637, 2023.

\bibitem[Xu et~al.(2018)Xu, Ouyang, Wang, and Sebe]{pad-net}
Dan Xu, Wanli Ouyang, Xiaogang Wang, and Nicu Sebe.
\newblock Pad-net: Multi-tasks guided prediction-and-distillation network for
  simultaneous depth estimation and scene parsing.
\newblock In \emph{CVPR}, pages 675--684, 2018.

\bibitem[Xu et~al.(2023)Xu, Li, Yuan, Yang, and Zhang]{mqt}
Yangyang Xu, Xiangtai Li, Haobo Yuan, Yibo Yang, and Lefei Zhang.
\newblock Multi-task learning with multi-query transformer for dense
  prediction.
\newblock \emph{IEEE TCSVT}, 2023.

\bibitem[Yang et~al.(2019)Yang, Liu, Chen, and Tong]{fedsurvey}
Qiang Yang, Yang Liu, Tianjian Chen, and Yongxin Tong.
\newblock Federated machine learning: Concept and applications.
\newblock \emph{ACM Trans. Intell. Syst. Technol.}, 10\penalty0 (2):\penalty0
  12:1--12:19, 2019.

\bibitem[Ye et~al.(2021)Ye, Lin, Yue, Guo, Xiao, and Zhang]{moml}
Feiyang Ye, Baijiong Lin, Zhixiong Yue, Pengxin Guo, Qiao Xiao, and Yu Zhang.
\newblock Multi-objective meta learning.
\newblock In \emph{NeurIPS}, pages 21338--21351, 2021.

\bibitem[Ye and Xu(2022)]{invpt}
Hanrong Ye and Dan Xu.
\newblock Inverted pyramid multi-task transformer for dense scene
  understanding.
\newblock In \emph{ECCV}, pages 514--530, 2022.

\bibitem[Ye and Xu(2023)]{invpt++}
Hanrong Ye and Dan Xu.
\newblock Invpt++: Inverted pyramid multi-task transformer for visual scene
  understanding.
\newblock \emph{arXiv preprint arXiv:2306.04842}, 2023.

\bibitem[Yu et~al.(2021)Yu, Zhang, Qin, Xu, Wang, Liu, Tian, and Chen]{fed2}
Fuxun Yu, Weishan Zhang, Zhuwei Qin, Zirui Xu, Di Wang, Chenchen Liu, Zhi Tian,
  and Xiang Chen.
\newblock Fed2: Feature-aligned federated learning.
\newblock In \emph{ACM SIGKDD}, pages 2066--2074, 2021.

\bibitem[Yu et~al.(2020)Yu, Kumar, Gupta, Levine, Hausman, and Finn]{pcgrad}
Tianhe Yu, Saurabh Kumar, Abhishek Gupta, Sergey Levine, Karol Hausman, and
  Chelsea Finn.
\newblock Gradient surgery for multi-task learning.
\newblock In \emph{NeurIPS}, 2020.

\bibitem[Yurochkin et~al.(2019)Yurochkin, Agarwal, Ghosh, Greenewald, Hoang,
  and Khazaeni]{pfnm}
Mikhail Yurochkin, Mayank Agarwal, Soumya Ghosh, Kristjan~H. Greenewald,
  Trong~Nghia Hoang, and Yasaman Khazaeni.
\newblock Bayesian nonparametric federated learning of neural networks.
\newblock In \emph{ICML}, pages 7252--7261, 2019.

\bibitem[Zhang et~al.(2021)Zhang, Zhou, Li, Cui, Xie, and Yang]{vpt1}
Xiaoya Zhang, Ling Zhou, Yong Li, Zhen Cui, Jin Xie, and Jian Yang.
\newblock Transfer vision patterns for multi-task pixel learning.
\newblock In \emph{ACM MM}, pages 97--106, 2021.

\bibitem[Zhang et~al.(2018)Zhang, Cui, Xu, Jie, Li, and Yang]{jtrl}
Zhenyu Zhang, Zhen Cui, Chunyan Xu, Zequn Jie, Xiang Li, and Jian Yang.
\newblock Joint task-recursive learning for semantic segmentation and depth
  estimation.
\newblock In \emph{ECCV}, pages 235--251, 2018.

\bibitem[Zhang et~al.(2019)Zhang, Cui, Xu, Yan, Sebe, and Yang]{pap-net}
Zhenyu Zhang, Zhen Cui, Chunyan Xu, Yan Yan, Nicu Sebe, and Jian Yang.
\newblock Pattern-affinitive propagation across depth, surface normal and
  semantic segmentation.
\newblock In \emph{CVPR}, pages 4106--4115, 2019.

\bibitem[Zhou et~al.(2020)Zhou, Cui, Xu, Zhang, Wang, Zhang, and Yang]{psd}
Ling Zhou, Zhen Cui, Chunyan Xu, Zhenyu Zhang, Chaoqun Wang, Tong Zhang, and
  Jian Yang.
\newblock Pattern-structure diffusion for multi-task learning.
\newblock In \emph{CVPR}, pages 4514--4523, 2020.

\bibitem[Zhu et~al.(2021{\natexlab{a}})Zhu, Lin, Lu, Lin, and Han]{commun3}
Ligeng Zhu, Hongzhou Lin, Yao Lu, Yujun Lin, and Song Han.
\newblock Delayed gradient averaging: Tolerate the communication latency for
  federated learning.
\newblock \emph{NeurIPS}, 34:\penalty0 29995--30007, 2021{\natexlab{a}}.

\bibitem[Zhu et~al.(2021{\natexlab{b}})Zhu, Hong, and Zhou]{fedgen}
Zhuangdi Zhu, Junyuan Hong, and Jiayu Zhou.
\newblock Data-free knowledge distillation for heterogeneous federated
  learning.
\newblock In \emph{{ICML}}, pages 12878--12889, 2021{\natexlab{b}}.

\bibitem[Zhuang et~al.(2023)Zhuang, Wen, Lyu, and Zhang]{MAS}
Weiming Zhuang, Yonggang Wen, Lingjuan Lyu, and Shuai Zhang.
\newblock Mas: Towards resource-efficient federated multiple-task learning.
\newblock In \emph{ICCV}, pages 23414--23424, 2023.

\end{thebibliography}
}

\end{document}